\documentclass[letterpaper, 10 pt, conference]{ieeeconf}  
\usepackage{graphicx}
\usepackage{amsmath}
\usepackage{booktabs}
\usepackage{hyperref}
\usepackage{soul}
\usepackage{xcolor}
\usepackage[dvipsnames]{xcolor}
\usepackage{tcolorbox}
\usepackage{graphicx}
\usepackage{caption}
\usepackage{balance}
\usepackage{cite}

\IEEEoverridecommandlockouts                              

\title{\LARGE \bf Pedicel-Targeted Strawberry Harvesting via Imitation Learning on a Low-Cost Teleoperated Robotic Arm\textcolor{NavyBlue}{\textbf{*}}}

\author{Nitin Rai$^{1}$, Won Suk Lee$^{1 \dagger}$, Hongyoung Jeon$^{2}$, Heping Zhu$^{2}$ \\ 
$\dagger$ Corresponding author (\tt wslee@ufl.edu)
\thanks{$^{1}$Department of Agricultural and Biological Engineering, University of Florida, Gainesville, FL 32611 USA.
        {\tt\small}}%
\thanks{$^{2}$United States Department of Agriculture, Agricultural Research Service, Application Technology Research Unit, Wooster, OH 44691 USA.
       {}}%
       \thanks{\textcolor{NavyBlue}{*This paper was peer-reviewed and accepted in the IEEE/RSJ International Conference on Intelligent Robots and Systems (IROS), ``From Innovation to Adoption: Hardware, AI, Policy, and Funding Challenges in Agricultural Robotics Workshop,'' Pittsburgh, Pennsylvania, 2026.}}
}

\begin{document}

\maketitle
\thispagestyle{empty}
\pagestyle{empty}

\begin{abstract}

Precise pedicel-targeted grasping in strawberry harvesting remains a critical bottleneck due to the pedicel's small diameter (1.3-3 mm), color similarity with surrounding foliage, and precise force control required for clean abscission zone detachment. As a market-grade crop, strawberry aesthetic value makes bruise-free pedicel detachment essential over direct fruit grasping. Classical approaches struggle to unify vision models, motion planning, and contact manipulation into a robust framework, particularly in occluded canopy environments. Therefore, this work presents a visuomotor imitation learning pipeline for pedicel-targeted strawberry harvesting that unifies perception, motion planning, and contact manipulation from collecting human teleoperation demonstrations, training an Action Chunking with Transformers (ACT) policy with a ResNet-50 visual backbone, and deploying the learned policy autonomously on a Jetson AGX Orin edge platform at 15 Hz. The trained policy achieved consistent loss convergence from 2.84 to 0.091 over 100K gradient steps on 130 demonstrations. We have also established a reproducible open-source pipeline with publicly released dataset and model weights for the agricultural robotics community.~Full dataset is available at~\href{https://huggingface.co/datasets/nitindominicrai/strawberry_pedicel_grasp_teleoperation}{\textcolor{blue}{nitindominicrai/strawberry\_pedicel\_grasp\_teleoperation}} and trained policy at~\href{https://huggingface.co/nitindominicrai/act_strawberry_pedicel}{\href{https://huggingface.co/nitindominicrai/act_strawberry_pedicel}{\textcolor{blue}{nitindominicrai/act\_strawberry\_pedicel}}}. To replicate this work, all code, configurations, and deployment instructions are publicly available at~\href{https://github.com/nitin-dominic/autonomous_strawberry_pedicel_harvesting}{\textcolor{blue}{autonomous\_strawberry\_pedicel\_harvesting}}.

\end{abstract}


\section{INTRODUCTION}

Strawberries \emph{(Fragaria $\times$ ananassa)} are widely grown across Florida during the winter season, typically from October to mid-March. In 2024, Florida planted 16,200 acres (6,556 ha) of strawberries and produced over 332M lbs (150.6M kg) of fruit \cite{b1}. Maintaining this level of production and fruit quality requires timely harvesting by February-March, placing significant demand on the available labor force. During peak harvesting season (Feb-March), an average of 110 workers per 100 acres are required, underscoring the urgent need for ground-based robotic harvesting robots in strawberry production~\cite{b_labor}. 

Over the past decade, significant research has focused on vision-guided robotic end-effectors for direct strawberry fruit grasping \cite{b_li2025,b_yang2025}. However, direct fruit contact can introduce bruising risks that degrade aesthetic quality and market value under uncontrolled grasping force \cite{b_chauhan2025}. Pedicel-targeted grasping offers a less damaging alternative, yet presents its own challenges, the pedicel measures only 1.3-3 mm in diameter, shares spectral similarity with surrounding foliage and calyx, and must be localized within a densely occluded canopy. Classical CV pipelines struggle with such slender structure detection, and coupling detection with downstream motion planning introduces compounding error in unstructured field conditions.


\begin{figure}
    \centering
    \includegraphics[width=1\linewidth]{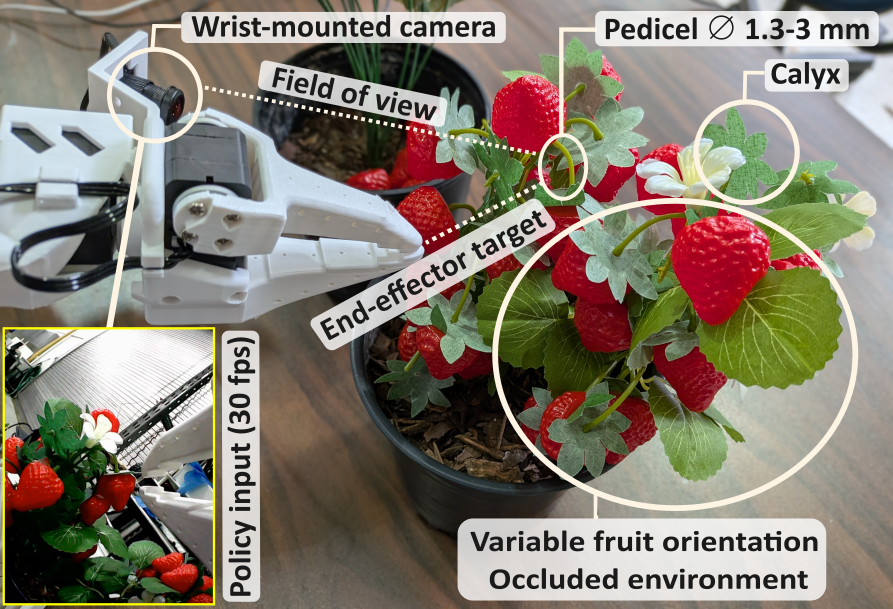}
    \caption{Pedicel-targeted strawberry harvesting using the SO-ARM101 6-DOF robotic arm. The task demands gripper alignment to a 1.3-3 mm pedicel under foliage occlusion and color ambiguity, addressed here as a unified visuomotor imitation learning problem via policy trained on human teleoperation demonstrations.}
    \label{fig1}
\end{figure}

Therefore, in this work, we present an \textit{end-to-end imitation learning approach for pedicel-targeted strawberry harvesting that bypasses explicit intermediate perception 
and planning stages entirely}. Using a low-cost teleoperation platform, we collect human demonstrations of the full reach-grasp-twist-pull sequence and train an Action Chunking with Transformers (ACT) policy directly from synchronized wrist-camera observations and joint-space actions~\cite{c_act}. The trained policy is deployed autonomously on a Jetson AGX Orin edge device, providing a fully self-contained harvesting system. The contributions of this work are as follows:
\begin{enumerate}
\item [a.] A teleoperation protocol and open-access dataset for strawberry pedicel grasping, comprising episodes of the complete reach-grasp-twist-pull~sequences recorded at 30 FPS.

\begin{figure*}[ht!]
    \centering
    \includegraphics[width=\textwidth]{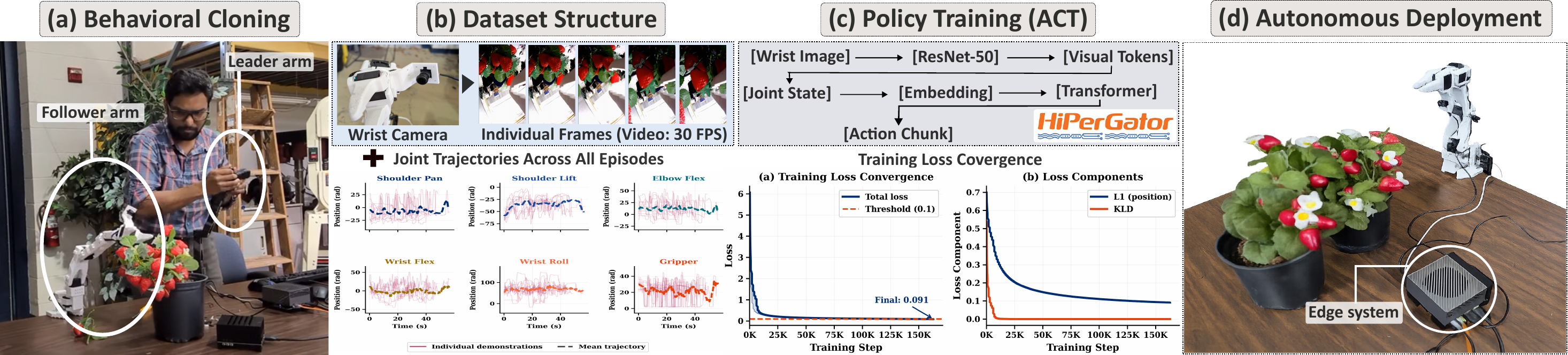}
    \caption{Behavioral cloning pipeline for pedicel-targeted strawberry harvesting. (a) A human operator collects reach-grasp-pull-retract demonstrations via bilateral teleoperation. (b) Each episode is stored as synchronized wrist-camera frames and 6-DOF joint positions (130 episodes, 250k frames); joint trajectories are shown with mean (dashed) (c) An ACT policy with ResNet-50 backbone is trained via behavioral cloning over 150K steps. (d) The policy runs autonomously on a Jetson AGX Orin at 15 Hz from live wrist-camera input.}
    \label{fig2}
\end{figure*}

\item [b.] Training and evaluation of an ACT visuomotor policy with ResNet-50 as a backbone on the collected dataset, demonstrating convergent learning behavior required for pedicel detachment.

\item [c.] Deployment and benchmarking of the trained policy on a Jetson AGX Orin, assessing the inference constraints and failure modes of edge-deployment.
\end{enumerate}

\section{METHODOLOGY}

Targeted pedicel detachment in strawberry harvesting presents a uniquely constrained manipulation problem at the intersection of twisting, pulling, and detaching in an unstructured scene perception.~Unlike strawberry grasping and picking, successful detachment requires multi-phase end-effector trajectory, including \textit{approach under partial-occlusion, sub-millimeter gripper alignment to the stem axis, controlled closure to avoid fruit bruising,} and \textit{pull-twist motion sufficient to exceed the detachment zone without slipping}.~To address these constraints, we propose a teleoperation-based behavioral cloning (imitation learning) framework that captures expert demonstration data encoding the full manipulation sequence, from approach to detachment, into a unified visuomotor dataset, which is subsequently used to train an end-to-end imitation learning policy (Fig.~\ref{fig2}).



\subsection{Data Collection via Human Demonstration}

We collected teleoperation demonstrations \textit{(n$\approx$130 episodes)} of strawberry pedicel-targeted grasping consisting of around 250k frames. Each episode consisted of a complete reach-approach-grasp sequence lasting 19-42 seconds, where the operator maneuvered the arm through dense foliage to align the end-effector with the pedicel and performed clutch-based grasping to simulate detachment contact behavior. However, the full abscission zone detachment was not performed due to the absence of a task-specific end-effector. The dataset collected focused on capturing the approach strategy and contact phase, which represent the primary perceptual and manipulation challenges. 


\subsection{Policy Training and Deployment}

We adopt the ACT framework as our visuomotor policy and deploy it on a low-cost SO-ARM101 6-DOF robotic arm \cite{c_act}. We trained an ACT policy with ResNet-50 as the visual backbone, replacing the standard ResNet-18 to improve feature extraction for the small-target pedicel detection sub-task. Hyperparameters were tuned for our data regime \emph{(n=130)}.~The trained ACT policy was deployed on a NVIDIA Jetson AGX Orin (Jetpack 5.1 with CUDA 11.4) at an average 25 Hz with FP32 precision mode.

\section{RESULTS AND DISCUSSION}
\subsection{Training Convergence}

The ACT policy was trained for 160k gradient steps on a single NVIDIA B200 GPU ($\approx$195 GB VRAM) using the collected demonstration dataset. Total loss decreased from an initial value of 2.84 to  (Fig.~\ref{fig2}b), with the KL divergence term stabilizing near zero around 17k steps, indicating convergence of the Conditional Variational Autoencoder (CVAE) latent distribution (Fig.~\ref{fig2}c). The ``l1'' joint prediction error plateaued at 0.062, demonstrating that the policy successfully learned the demonstrated reach-grasp-pull-retract trajectories. 

\subsection{Deployment Characteristics}

\begin{figure}[ht!]
    \includegraphics[width=0.5\textwidth]{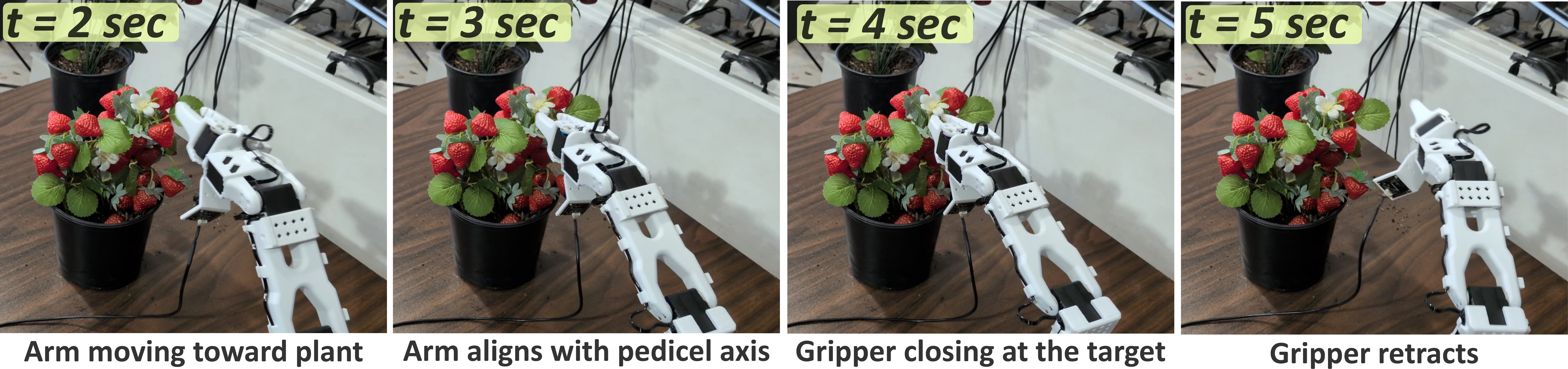}
    \caption{Camera keyframes during autonomous pedicel-targeted harvesting inference. The policy executes a reach-align-contact sequence learned from teleoperation demonstrations, without human intervention.}
    \label{fig3}
\end{figure}

The trained policy was deployed on a Jetson AGX Orin at $\approx$15-17 Hz using FP32 inference with a ResNet-50 visual backbone. The arm consistently reproduced the learned approach trajectory toward the strawberry plant, confirming successful visuomotor policy transfer from human demonstrations to autonomous execution (Fig.~\ref{fig3}). 

Although imitation learning provides an end-to-end pipeline for pedicel-targeted grasping, several challenges remain open for future investigation: 

\begin{enumerate}
    \item [a.] Short pedicels connecting directly to the main stem constrain end-effector maneuverability, requiring more compact gripper designs.
    \item [b.] Wrist camera placement is very critical as suboptimal mounting can occlude the gripper-pedicel contact region, degrading policy performance.
    \item [c.] Sim-to-real transfer is challenging for pedicel grasping, as synthetic environments fail to replicate the dynamic nature of the strawberry pedicel and the light variation within a real plant canopy, limiting the utility of large-scale data augmentation strategies common in general-purpose imitation learning.
    \item [d.] A cutting-based end-effector requires sub-millimeter precision to clip individual pedicels without contacting the calyx or fruit surface. 
\end{enumerate}

\section{CONCLUSIONS}

We presented an end-to-end imitation learning pipeline for pedicel-targeted strawberry harvesting on a low-cost teleoperation platform. The ACT policy successfully learned the approach and grasp trajectory from 130+ human demonstrations and was deployed autonomously on a Jetson Orin AGX at $\approx$15 Hz. While full pedicel detachment was not reliably achieved in this initial system, the pipeline
validates the core technical approach and identifies concrete directions for improvement.

Our future work will focus on three areas. First, expanding the demonstration dataset to at least 300 episodes with targeted coverage of varied berry positions and approach angles is
expected to substantially improve grasp reliability, as behavioral cloning policies are known to scale well with demonstration diversity~\cite{hussein2017imitation}. Second, we plan to integrate a phase-aware weighted loss function within ACT-policy training that assigns higher importance to the critical contact and detachment frames, addressing the precision asymmetry between the approach and grasp phases. Third, upgrading the Jetson firmware to JetPack 6.2 will enable 30 Hz inference via NVIDIA's prebuilt PyTorch wheels, eliminating the current speed constraint and enabling real-time reactive control at the pedicel scale.


\addtolength{\textheight}{-12cm}   





\section*{ACKNOWLEDGMENT}

This work was supported by USDA-ARS in Wooster, Ohio (Agreement \#:~58-5082-4-040).~Mention of company or trade names is for description only and does not imply endorsement by the USDA. The USDA is an equal opportunity provider and employer.




\balance

\end{document}